\documentclass{article} 
\usepackage{iclr2020_conference,times}

\usepackage{booktabs, multirow} 
\usepackage{soul}
\usepackage{changepage,threeparttable} 

\usepackage{amsmath,amsfonts,bm}
\usepackage{physics}

\def\eqref#1{equation~\ref{#1}}

\def\1{\bm{1}}

\def\rvc{{\mathbf{c}}}

\DeclareMathAlphabet{\mathsfit}{\encodingdefault}{\sfdefault}{m}{sl}
\SetMathAlphabet{\mathsfit}{bold}{\encodingdefault}{\sfdefault}{bx}{n}

\newcommand{\R}{\mathbb{R}}

\usepackage{graphicx}
\usepackage{hyperref}
\usepackage{url}
\iclrfinaltrue

\title{Learning and Optimization of Blackbox Combinatorial Solvers in Neural Networks}

\author{T.J. Wilder \\
School of Computer Science\\
University of Wisconsin - Madison\\
Madison, WI 53703, USA \\
\texttt{tj@wilder.tj}
}

\begin{document}

\maketitle

\begin{abstract}
The use of blackbox solvers inside neural networks is a relatively new area which aims to improve neural network performance by including proven, efficient solvers for complex problems. Existing work has created methods for learning networks with these solvers as components while treating them as a blackbox. This work attempts to improve upon existing techniques by optimizing not only over the primary loss function, but also over the performance of the solver itself by using Time-cost Regularization. Additionally, we propose a method to learn blackbox parameters such as which blackbox solver to use or the heuristic function for a particular solver. We do this by introducing the idea of a hyper-blackbox which is a blackbox around one or more internal blackboxes.
\end{abstract}

\section{Introduction}
In computer science, neural networks continue to be more and more widely used. They can be used to solve many problems in a highly general way, allowing these problems to be dealt with primarily by using appropriate architecture and sufficient data. On the other hand, there are classical algorithmic techniques, such as graph algorithms and SAT-solvers, which are highly optimized and studied. However, rather than being highly general, they are usually very specific to their exact problem and feature space. Since there are so many different problems and problem domains, the goal then is to combine these two approaches in an efficient and effective way.

By fusing these two techniques together, it becomes possible to leverage the incredible flexibility and data-based feature extraction of neural networks while also taking advantage of highly optimized algorithmic implementations of various problems. In general, this means that it becomes possible to learn a model with arbitrary, but related, input and output domains that can still take advantage of highly specific combinatorial and algorithmic solvers as a component of the model architecture.

The primary issue with introducing arbitrary algorithmic functions in the context of neural networks is that arbitrary functions are generally not differentiable. While not an issue for solving a known problem, this poses a large problem for learning new problems as neural networks typically depend on gradient descent based algorithms for learning and optimization. Despite gradient descent being the primary method for learning, there are some other techniques which may be applicable for learning without a gradient such as evolutionary models (\cite{10.1162/106365602320169811}) and closed form minimization (\cite{taylor2016training}). Similar to a gradient based approach however, it would be non-trivial to adapt combinatorial solvers to fit into those architectures as well. 

There are also techniques which focus on the problem from the opposite side. Rather than learning without gradients, they could focus on relaxing, altering, or approximating the objective function or of a baseline algorithm such that it has a usable gradient. However, doing this leads to sub-optimal results in terms of output correctness, but can also be worse in terms of performance (\cite{10.1145/1374376.1374414}). To circumvent these issues, \citeauthor{vlastelica2019differentiation} introduced a method of differentiating combinatorial solvers by treating them as a blackbox and implementing a special backward pass in order to compute a reasonable gradient. This approach improves on previous results in that it solves the exact problem using any solver which optimizes a linear objective function and generates a meaningful gradient. It does however have some limitations beyond just minimizing a linear objective function. One is that it guarantees optimality of solution, but only promises high performance in that the solver itself is usually bounded in complexity. Two is that it requires exact knowledge of which solver to use and how it will fit into the architecture (\cite{vlastelica2019differentiation}). There are more, but addressing those limitations two is the objective of this work.

In this paper, we focus on overcoming those two limitations in a way that both maximizes the advantages of neural networks and circumvents the limitations of a single blackbox combinatorial solver in two novel ways. The first way is novel, not in complexity, but in that it is not useful in traditional neural networks. This first technique, we call \textit{Time-cost Regularization}. Similar to classic regularization techniques which effects the weights such as $\ell_1$-regularization or more specific techniques like ReLU Stability loss, time-cost regularization adds a penalty based on the time cost of blackbox components of the neural network. It will be justified and explained more in our Methods section, but the intuition is that fixed neural networks have a roughly constant amount of computation when executed in the same computational environment, but algorithmic solvers with the same \textit{solution} may not take the same amount of computations to find that solution. The second technique is related to the first technique and addresses the issue of requiring knowledge of the exact solver and architecture beforehand. This technique uses a hyper-blackbox and provides additional, learned hyperparameters to that hyper-blackbox such that it can choose the optimal solver. In this work, we focus primarily on choosing a solver such that it minimizes the time-cost of the problem. We also discuss ramifications for accuracy by optimizing over multiple objectives.

\section{Related Work}
As noted above, the primary work that this builds on is that of Vlasteluca et al. They built their work on \cite{br2017optnet} and \cite{NIPS2010_4069}, among others. They introduced a method for including blackbox combinatorial solvers in neural networks by creating a novel backward pass algorithm which provides an informative gradient despite the solver itself not being differentiable. The method is complex to understand and prove and has few explicit requirements that must be met by the solver. It must map continuous inputs to discrete outputs and it must minimize a linear cost function. These constraints are related to the informativeness of the computed gradient. Instead of explicitly requiring discrete outputs, we believe that the more accurate requirement is that loss function induced by that output should be discrete for a given network state. A discrete loss helps justify the linearization of the loss function later while a non-discrete loss function would require more analysis to prove the informativeness of the gradient. This could be a discrete set of real numbers too. For example, a shortest path algorithm will produce a path, so the discrete loss function could the set of losses comparing all possible paths to the correct path. But a discrete output necessitates a discrete loss function because the loss function has discrete inputs. Besides those constraints, it places no assumptions on the set of constraints for the solver or on the structure of the output space.

For clarity, we use the same formulation and notation as described in their paper. The combinatorial solver receives continuous input $w \in W \subseteq \R^N$ and returns discrete output $y \in Y$ where $Y$ is a finite set of possible results. The function then minimizes the linear cost function $\rvc(w,y) = w \cdot \phi(y)$. The solver is then the function which minimizes that cost function. That is
\[
w \rightarrow y(w) \,\,\,\,\text{ s.t. }\,\,\,\, y(w) = \underset{y \in Y}{\arg\min} \,\,\rvc(w, y)
\]
Beyond their formulation of the blackbox solver itself, the other most relevant detail is their novel backward pass. This requires a few more components, presented here with minimal explanation. For a neural network with loss function $L$, they define a linearization $f(y)$ of that loss function at a specific point $\hat{y}$. Further, they define $f_\lambda(y)$ as the continuous interpolation of that linearized loss function. To define that, we need to define $y_\lambda(w)$ as the solution to a perturbed optimization problem. Formally, this gives us
\[
f(y) = L(\hat{y}) + \dv{L}{y}{}(\hat{y}) \cdot (y - \hat{y})
\]
\[
y_\lambda(w) = \underset{y \in Y}{\arg\min}\{\rvc(w, y) + \lambda f(y)\}
\]
Alternatively, we can devise a perturbed weight value $w'$ and define $y_\lambda$ equivalently using that
\[
y_\lambda(w) = \underset{y \in Y}{\arg\min}\left\{\rvc\left(w', y\right)\right\} \text{ where } w' = \hat{w} + \lambda \dv{L}{y}{} (\hat{y})
\]
\[
f_\lambda(w) = f(y_\lambda(w)) - \frac{1}{\lambda} \left[ \rvc(w, y(w)) - \rvc(w, y_\lambda(w)) \right]
\]
\[
\nabla f_\lambda(w) = -\frac{1}{\lambda} \left[y(w) - y_\lambda(w)\right]
\]
That last function, the gradient of $f_\lambda$ returns the perceived gradient of the blackbox solver. We can use the original output $y(w)$ from the forward pass for a start. It's relatively cheap to calculate the rest, but we do need to calculate the perturbed weights $w'$ from the forward pass values and loss and then run the blackbox solver on those perturbed weights, so the cost of the backward pass is actually slightly more than the cost of the forward pass which only has to run the solver.

The primary thing we've glossed over here is the hyperparameter $\lambda$. $\lambda$ represents a trade-off between the informativeness of the gradient and the faithfulness of the interpolation to the original function. There are detailed descriptions and guidelines in the paper as to what $\lambda$ means and what good values might be \cite{vlastelica2019differentiation}.

To demonstrate and describe the techniques contributed by this paper, we'll use a re-framing of the Shortest Path problem used by Vlasteluca et al. The Shortest Path problem in general takes as input a weighted, directed graph with a starting $S$ and ending vertex $E$ and returns an ordered list of edges which describes the path from $S$ to $E$ with the lowest total weight cost. Here, we'll consider a simplification where every vertex on the graph is cell on a 2D-grid and the edges to a cell from all horizontal and diagonally neighbors have the same positive weight. Thus we can simplify the input to a grid with a weight cost to enter each cell. We can also simplify the output to a set of cells that occur on the path because the order doesn't matter and path with repeats could never be lower cost than one with no repeats as long as the weights are non-negative.

Their paper only examined Dijkstra's Shortest Path algorithm, but we'll also compare it with A*. In order to do that though, it's worth taking a look at the differences between the two algorithms first. Firstly, an essential detail is that they both produce the same optimal path when given the same input. So with all other things being equal, including one or the other inside a neural network will give you equivalent optimal results. And because of the equivalent paths, the neural network updates will be identical. So the question then is why we could care about their differences. In general, the big difference between them is the use of a heuristic function in A* which attempts to better guide the algorithm towards the optimal path. We can describe the choice of heuristic function as a hyperparameter on the algorithm which, so long as it's admissible, only effects the efficiency and run-time of the algorithm and not the optimality (\cite{reddy2013path}). Because of this heuristic function, it seems like A* should be faster, but that actually depends on the heuristic function. The A* algorithm can actually be more expensive as it needs to update the costs of nodes dynamically, while Dijkstra guarantees that every time it sees a node it will along the minimum cost for that path. Thus, if the heuristic is not very informative or if the shortest path is very simple, A* can actually be more expensive overall, despite the perceived optimizations. Optimizing the choice of shortest path algorithm and heuristic functions is a driving force in the rest of the paper.

\section{Methods}
\subsection{Time-cost Regularization}

Let us first formalize the idea of a \textit{time-cost regularization} and then motivate why it is helpful and address possible issues. Time-cost regularization can be simply defined by adding on a regularization term onto the existing loss function. The following shows a simple optimization function with time-cost regularization and $\ell$-1 regularization. Let $\alpha$ and $\beta$ be hyperparameters, $W$ be the vector of all weights, $f_B(x)$ be the blackbox solver portion of $f(x)$, and $t(f_B(x))$ be the duration in seconds of the blackbox solver process, then
\[
L(x, y) = (y - f(x))^2 + \alpha ||W||_1 + \lambda_t t(f_B(x))
\]
In intuitive terms, minimizing the total loss also includes minimizing the time spent by the blackbox solver. In ordinary neural networks, there's no sense in optimizing by time spent directly since the time spent doing neural network calculations will stay constant across changes in weights. Instead we'd normally choose to try to optimize the number of neurons or layers in order to minimize the overall function time. In this case however, the important difference is that a blackbox solver can find the same path in a faster or slower time depending on the exact inputs that get passed in.

Let's go through a simple, illustrative example to see why this is the case. Let's assume that our shortest path along a grid is directly along the diagonal from the top-leftmost corner to the bottom-rightmost corner. Now assume we have two different weight matrices with that same shortest path. The first holds all weights constant so the shortest path is the one with the fewest steps directly along the diagonal. The second matrix has weights of $0.1$ along the diagonal, and $10,000$ everywhere else. It's trivial to show that these both have the same shortest path. So instead lets focus on the grid cells visited by Dijkstra's algorithm for each of these examples.

For the first case, Dijkstra will continue inspecting all adjacent cells starting from the first corner because all the cells have the same weight. The amount of cells will thus spread out in all directions, eventually visiting (or at least inspecting) every single node before arriving at the goal position. In the second case, Dijkstra will instead only inspect the elements along or adjacent to the shortest path, because it visits them in minimal-cost order. Unless it was a square grid that was at least $100,000$ cells wide, the cost of trying to along the diagonal would always be less than the cost of visiting any of the $10,000$ cost cells. So for these cases, with a $n \times n$ grid, the first one requires visiting $n^2$ cells and the second one requires visiting $n$ cells.

So while that was a bit of a pathological example, it should not be surprising that more subtle changes in the grid can also cause more subtle changes in the time-cost of finding the shortest path. The same goes with weights in the A* case, but in that case, the heuristic function can also have a significant impact on performance. In fact, for the case where the heuristic always returns 0, A* degenerates into exactly Dijkstra's algorithm. We'll give evidence that this works in the Experiments section, but this should show that it's possible to optimize for time when our primary goal is simply shortest path.

It's great to show that it's possible, but the question remains whether it's a good idea to optimize for time. At an intuitive level, optimizing for time in this case is basically encouraging the neural network to try to solve the problem for itself. If it can do that, such as by producing the second case we discussed just now, then the blackbox solver can trivially find the shortest path since the neural network has already described it in the "time-optimal" way. If it can't solve the problem itself, then all it needs to do is represent the grid well enough that the blackbox solver can solve it, even if that takes a lot of time. This trade-off is naturally explained by the time-cost constant ($\lambda_t$ in the above loss equation) which can scale the time-cost relative to the accuracy and any other regularization parameters. A well-chosen constant should nudge the learning towards learning grid weights which are easy to solve. Though a constant that's too high could prioritize speed over correctness and only learn to create our case 2 examples regardless of the input simply because it's fast to solve. Somewhere in the middle, and the network may tend to create grid weights which generate similar cost paths which are much easier to calculate. One example might be that walking over a mountain may be a more expensive direct path, but finding a path around a large mountain range may yield only a slightly better path but could take a much longer time to optimize around.

In this case the choice of hyperparameters, like for many other trade-offs in machine learning, is heavily reliant on the problem space and requirements for speed, correctness, and optimality. If the goal is purely to find optimal solutions, it's unlikely that time-cost regularization will be of significant help, though a small value may still lead to faster run-time with a minimal loss in accuracy. In our Experiments section, the difference in time between batches is on the order of tenths of seconds for a simple 12x12 grid, but for more complex examples with a larger problem space or a higher time complexity algorithm, then the difference could be far more significant. However, in this case, at least it is intuitive how the trade-off works since it's a lot easier to equate time and accuracy than something more arbitrary than traditionally regularizing based on weights. For shortest path, you might have an error of 2 for each step which doesn't match the expected, which would be equivalent to spending an extra second on the blackbox at $\lambda_t = 2$.

Ironically, the method of blackbox differentiation that we're using is only proven to work for blackboxes which minimize a cost function. So you can't trade-off the optimality of the solver itself, but through using things like time-cost regularization you can indirectly trade-off the optimality of the whole network.

\subsection{Hyper-Blackbox}
The idea behind a hyper-blackbox is to define a blackbox which contains multiple internal blackboxes. On its own, that means that the hyper-blackbox can effectively delegate its tasks to the internal blackboxes. By constructing a hyper-blackbox, we can put that into a neural network instead of putting in multiple separate blackboxes. There's an important requirement to be able to combine these solvers like this. But it's actually the same as our requirement to include a single blackbox in that the hyper-blackbox must always optimize the same cost function. This does not require that the internal blackboxes take the same input or produce identical output however. It only requires that there exists a mapping from any given input of the hyper-blackbox to at least one internal blackbox and there exists a mapping from the output of any of the internal blackboxes to the minimal-cost output required of the hyper-blackbox. Differentiating over the hyper-blackbox can either take the same form as differentiating over the normal blackboxes, or you can treat it as a simpler component by only using that technique on the internal blackboxes. This primarily depends on whether you're doing any non-differentiable calculations to determine which internal blackbox to use.

Intuitively, it's not surprising that you could combine two algorithms with the same inputs and outputs like Dijkstra and A*, but these mappings also give you some flexibility in the types of algorithms you can use internally. For instance, perhaps your network generates one grid but needs to calculate the shortest paths between multiple pairs of points. You could use a copy of Dijkstra or A* for each pair of points, or you could instead use something like the Floyd-Warshall all pair shortest path algorithm which computes all the shortest paths at once. Using a hyper-blackbox containing both methods gives some flexibility for better optimizing the time-cost. The question is then how to take advantage of this flexibility.

The first way is to include some decision process inside the hyper-blackbox itself. As in our Experiments section, here we'll refer to the case where we have a hyper-blackbox which contains only Dijkstra and A*. As addressed earlier, in order for A* to be more efficient than Dijkstra, it requires an admissible, informative heuristic function. One simple decision process could be a scan of the weights of the input matrix to determine if our heuristic function is informative. In the case of a grid of arbitrary non-negative weights, it's difficult to make a constant heuristic since its admissibility requires it to always underestimate the cost of the remaining path from any given point. So we can instead considered the per-grid heuristic $max\_steps\_to\_goal \times min\_weight$. It's possible to do better by doing more calculations in the decision process, but in this case we can check if the minimum weight is relatively close to other weights. If it's not close to other weights then it will massively underestimate the path cost at every point and won't be informative. If it is close to other weights, then it may be a reasonable lower bound.

The second way, which could be used alongside the first way, is to add additional parameters to the hyper-blackbox. This could be as simple as a single "choice" parameter, but could be more complicated inputs or decision metrics that could get calculated by the neural network alongside the grid weights. In the case of a hyper-blackbox containing Dijkstra and A*, we can use a single choice parameter where a value of $\geq 0.5$ means we use A* and a value of $< 0.5$ means we use Dijkstra. The mapping function from the hyper-blackbox input to the internal blackbox input simply removes that choice parameter and passes the remaining parameters as weights to the chosen internal blackbox. In this case, there's no need to do additional mapping on the output. Instead of encoding separate choice parameters, we may choose to add a skip connection from our input layer directly down to the hyper-blackbox and determine which internal blackbox to use based on the input layers. One trivial way this could be used is to directly encode the choice of solver as one of the inputs to the neural network.

The choice parameter here is quite literally the neural network attempting to make a choice for which algorithm it should minimize the loss. However, because the default formulation should have no difference in loss between two optimal algorithms, in order to learn the choice parameter, we must introduce some additional information such as our time-cost regularization. Using time-cost regularization with a hyper-blackbox gives the neural network an additional way to reduce the run-time of the network. Instead of partially solving the problem itself, we also give the option for the network to choose an algorithm which it knows will be able to solve the problem faster. In theory, this also means that the network could learn to optimize the weights for either algorithm and can learn to pick the one for which it can better optimize those grid weights.

Another use for these additional parameters to the hyper-blackbox is to use them as hyperparameters for the solvers themselves. For instance, in addition to using a choice parameter for the hyper-blackbox, we could also add a heuristic parameter that A* could use to determine what heuristic to use as it finds the shortest path. This is just another way that the neural network can help the solver by providing more information without needing to solve the entire problem on its own.

\section{Experiments}
We tested the effects of time-cost regularization on the shortest path problem and neural architecture described by \citeauthor{vlastelica2019differentiation} on the 12x12 Warcraft Shortest Path dataset SP(12). In this problem, we train a neural network containing a blackbox shortest path solver to predict the shortest path when given an image. The image is passed through a convolutional neural network which outputs a $k \times k$ grid of weights. The weights are then passed into the blackbox shortest path solver and the loss is calculated based on the Hamming distance between the true and expected paths.

We consider three different blackbox solvers. The first is Dijkstra as used in the existing work. The second is A* using a heuristic of $max\_steps\_to\_goal \times min\_weight$. The third is a hyper-blackbox which contains both of the other blackboxes as internal blackboxes and has an additional choice parameter which the neural network calculates as additional input to the hyper-blackbox. To facilitate this additional choice parameter, we add an additional fully connected layer onto the existing convolutional neural network output for all three solvers. In the case of Dijkstra and A*, this layer is initialized as a $144 \times 144$ identity matrix (as a $12 \times 12$ has 144 parameters), while in the hyper-blackbox, it is a near-identity in that it is $145 \times 144$ containing a $144 \times 144$ identity matrix and a single row of small weights intended to cause the choice parameter to be near the decision boundary. For this hyper-blackbox, we actually treat it as a simple model instead of a full blackbox for simplicity. The internal blackboxes are still fully treated as blackboxes.

\begin{figure}[t]
    \centering
    \begin{tabular}{cc}
       \includegraphics[scale=0.30]{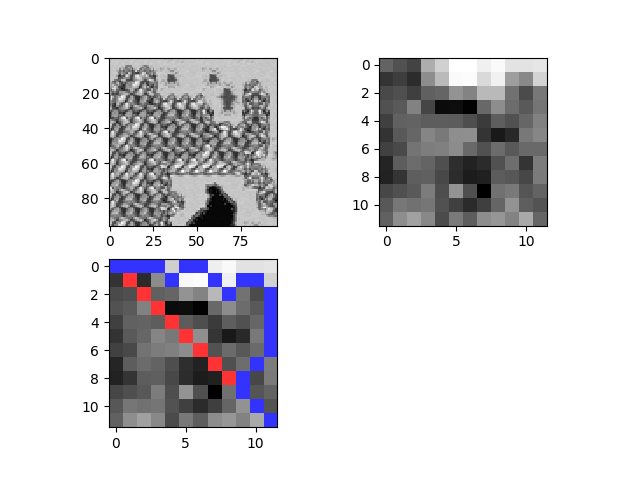}
       &  \includegraphics[scale=0.30]{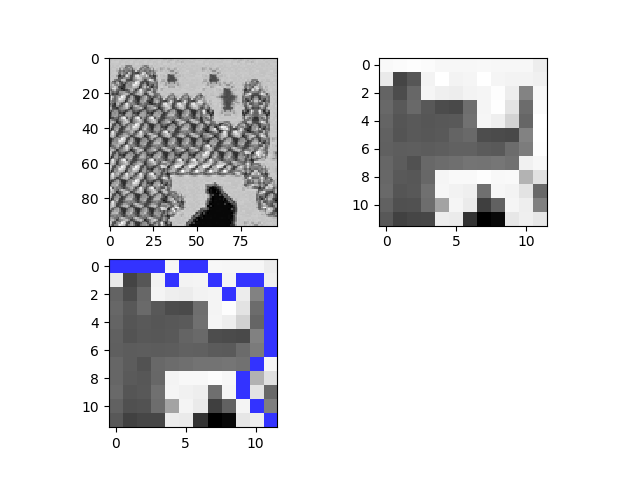}
    \\
       \includegraphics[scale=0.375]{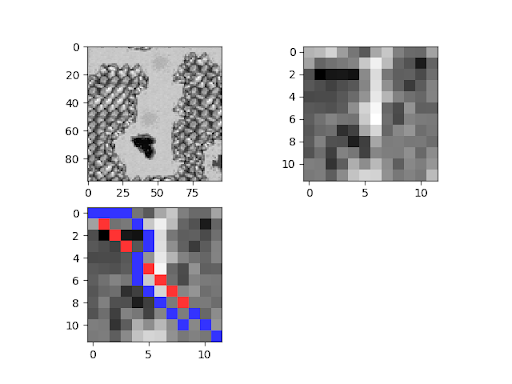}
       &  \includegraphics[scale=0.30]{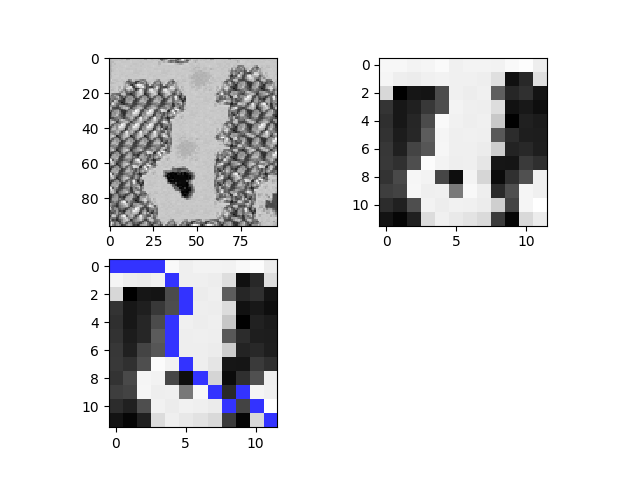}
    \end{tabular}
    \caption{Two pairs of matching sets of images. Each set of three images contains the original image, the calculated weight matrix, and the weight matrix overlaid with the optimal blue and predicted red path. The first column shows the results after 1 epoch and the second column shows the results after 5 epochs. When there is no red, that means that the optimal path was predicted exactly.}
    \label{fig:path_weights}
\end{figure}

Instead of observing blackbox solver time, we instead focus on the more general but less informative average batch time. We use this simply because it requires less changes to the code architecture. One thing that sticks out in the analysis is that the average batch time actually decreases regardless of using time-cost regularization. By examination of several weight matrices at different points in training, it seems apparent that the early weight matrices are only slightly better than random, so even though we can get decent shortest paths from them, it takes a long time to compute as the differences between weights for different terrain features are very similar. As the training goes on however, the features are better recognized and the difference in weights becomes much clearer. You can see this difference in Figure \ref{fig:path_weights}. You can see that in the later epochs, the white spots have much less cost so the algorithm can focus on exploring those. In the earlier examples, the weights are very similar so the algorithm needs to do more exploring to find the optimal path.

For all examples we used a $\lambda_t$ parameter of 50 for the time-cost regularization. The data in Table \ref{tab:acc} and Table \ref{tab:abt} refer to a single trial of each algorithm recording the average batch times and test accuracies of each one. We see that that the accuracy of each model is basically unaffected by the time-cost regularization (TCR). The only impact we see on accuracy is the introduction of $\ell$-1 regularization alongside TCR.

The impact on time is actually a lot less clear. Surprisingly, the average time using A* algorithm actually decreased faster without TCR, though not by an enormous amount. In contrast, the average time using Dijkstra's algorithm stayed fairly similar across epochs but was clearly lower when including TCR. In our hyper-blackbox case, the time started near time of A* and ended near the time of Dijkstra which, as shown in Table \ref{tab:ratio}, is actually because the model's initial parameters primarily chose A*, but the algorithm eventually learned to use exclusively Dijkstra because it was faster. This also makes sense based on their relative speed in Table \ref{tab:abt}.

Before we explore the strange A* anomaly a little bit, there's one more thing to point out with the hyper-blackbox version. Not only did it reach roughly the same maximum accuracy as any other model, but it and Dijkstra with TCR and $\ell$-1 were the only models to increase in accuracy between the 15th and 20th epochs. Though it's certainly a limited sample space, this may imply that a model that learned to optimize time across both A* and Dijkstra algorithms actually has a better overall understanding of the problem space and is thus better able to generate good weight matrices. The fact that the accuracy continued increasing could mean that it's also less susceptible to overfitting because it optimized over a larger space. Though it could also mean that the model was simply harder to learn and didn't generalize better and we simply didn't see it overfit within 20 epochs.

\begin{table}[!htp]\centering
\caption{Accuracy by Method and Epoch}\label{tab:acc}
\scriptsize
\begin{tabular}{lrrrrrrr}\toprule
Epoch &A* &A* w/ TCR &Dijkstra &Dijkstra w/ TCR &Dijkstra w/ TCR + $\ell$-1 & Hyper w/ TCR \\\cmidrule{1-7}
0 &0.8789 &0.8789 &0.8789 &0.8789 &0.8789 &0.8775 \\\cmidrule{1-7}
5 &0.9740 &0.9740 &0.9730 &0.9730 &0.9730 &0.9734 \\\cmidrule{1-7}
10 &0.9703 &0.9703 &0.9695 &0.9695 &0.9694 &0.9734 \\\cmidrule{1-7}
15 &0.9754 &0.9754 &0.9721 &0.9722 &0.9696 &0.9730 \\\cmidrule{1-7}
20 &0.9739 &0.9739 &0.9719 &0.9719 &0.9704 &0.9750 \\
\bottomrule
\end{tabular}
\end{table}

\begin{table}[!htp]\centering
\caption{Average Batch Time by Method and Epoch}\label{tab:abt}
\scriptsize
\begin{tabular}{lrrrrrrr}\toprule
Epoch &A* &A* w/ TCR &Dijkstra &Dijkstra w/ TCR &Dijkstra w/ TCR + L1 &Hyper w/ TCR \\\cmidrule{1-7}
0 &1.779 &1.773 &1.553 &1.544 &1.549 &1.781 \\\cmidrule{1-7}
5 &1.751 &1.762 &1.572 &1.597 &1.564 &1.626 \\\cmidrule{1-7}
10 &1.721 &1.743 &1.563 &1.561 &1.581 &1.601 \\\cmidrule{1-7}
15 &1.709 &1.721 &1.575 &1.561 &1.576 &1.593 \\\cmidrule{1-7}
20 &1.684 &1.700 &1.574 &1.560 &1.578 &1.586 \\
\bottomrule
\end{tabular}
\end{table}

\begin{table}[!htp]\centering
\caption{Proportion of A* Usage / Dijkstra Usage}\label{tab:ratio}
\scriptsize
\begin{tabular}{lr}\toprule
Epoch & Ratio (\#A* / \#Dijkstra) \\\cmidrule{1-2}
1 & 7.2 \\
2 & 1.9 \\
3 & 0.89 \\
4 & 0.48 \\
5 & 0.27 \\
6 & 0.16 \\
7 & 0.093 \\
8 & 0.054 \\
9 & 0.026 \\
10 & 0.013 \\
11 & 0.006 \\
12 & 0.002 \\
13 & 0.001 \\ 
14+ & 0.000 \\
\bottomrule
\end{tabular}
\end{table}

Now let's try to gain some insight into why A* got so much faster even without TCR. The likely reason is because of the heuristic in A*. Although the heuristic isn't being passed in explicitly as a parameter, because we're basing the heuristic on the minimum weight, it's actually being done implicitly. Without examining the heuristics directly it's difficult to say for sure how they changed across epochs, but it would make sense that adjusting the heuristic would have a more significant effect than TCR, especially since a more accurate set of weights will naturally produce a heuristic which is more informative on that problem. One factor of this is that none of the algorithms care about the weight at the starting location because it's always on the path except that for A*, that weight could be the minimum value and effect the heuristic of the whole function. It's also possible that TCR would have been significant if the model was given more epochs to learn and optimize. In order to fully understand this though, we'd have to run more trials and likely inspect the weights or heuristics as well, so we leave that to future work.

\section{Conclusion and Future Work}
We introduced novel techniques to better optimize the performance of blackbox solvers inside neural networks. We showed that the use of time-cost regularization and hyper-blackbox solvers can reduce the time cost of blackbox solvers without significant negative impacts on the accuracy of the model. More trials need to be run to see how significant the effects are relative to other factors. There are likely many further optimizations to be done in this area, but this is one step in that direction. 

For future work, it would be interesting to explore the differences in weight matrices produced by the introducing time-cost regularization. It could be that time-cost regularization hurts generality of the generated weight matrix for other tasks by only optimizing for the specific case of shortest path. In theory, that same learned function from pixels to weights could be used for other tasks, so exploring those could be very interesting. Additionally, we could look at the idea of only-optimal or near-optimal time-cost in that we scale the time cost based on how accurate the model already is. The idea then is that a model which has very low accuracy needs to focus on learning the weights, but once it's highly accurate, or perhaps perfect, then the time-cost regularization will help it learn weights that are easy to optimize over.

\bibliography{iclr2020_conference}
\bibliographystyle{iclr2020_conference}

\end{document}